\documentclass[runningheads]{llncs}

\usepackage{eccv}

\usepackage{eccvabbrv}

\usepackage{graphicx}
\usepackage{booktabs}
\usepackage{comment, multirow, array, algorithm, algpseudocode, amsmath, bbm}
\usepackage{makecell}
\usepackage{bbold}
\usepackage{amssymb}

\usepackage[accsupp]{axessibility}

\usepackage{hyperref}

\usepackage{orcidlink}

\begin{document}

\newcolumntype{L}[1]{>{\raggedright\arraybackslash}p{#1}}
\newcolumntype{C}[1]{>{\centering\arraybackslash}p{#1}}
\newcolumntype{R}[1]{>{\raggedleft\arraybackslash}p{#1}}
\newcolumntype{+}{>{\global\let\currentrowstyle\relax}}
\newcolumntype{^}{>{\currentrowstyle}}

\newcommand{\RomNum}[1]{\MakeUppercase{\romannumeral #1}}
\newcommand{\black}{\color{black}}
\newcommand{\red}{\color{red}}


\title{Forbes: Face Obfuscation Rendering via Backpropagation Refinement Scheme}

\titlerunning{Forbes}

\author{Jintae Kim\inst{1}\orcidlink{0009-0005-6873-5948} \and
Seungwon Yang\inst{2} \and
Seong-Gyun Jeong\inst{2} \and
Chang-Su Kim\inst{1}\orcidlink{0000-0002-4276-1831}}

\authorrunning{J. Kim et al.}

\institute{School of Electrical Engineering, Korea University, Seoul, Korea\and
42dot Inc.\\
\email{jtkim@mcl.korea.ac.kr, seungwon.yang@42dot.ai, seonggyun.jeong@42dot.ai, changsukim@korea.ac.kr}}

\maketitle

\begin{abstract}
A novel algorithm for face obfuscation, called Forbes, which aims to obfuscate facial appearance recognizable by humans but preserve the identity and attributes decipherable by machines, is proposed in this paper. Forbes first applies multiple obfuscating transformations with random parameters to an image to remove the identity information distinguishable by humans. Then, it optimizes the parameters to make the transformed image decipherable by machines based on the backpropagation refinement scheme. Finally, it renders an obfuscated image by applying the transformations with the optimized parameters. Experimental results on various datasets demonstrate that Forbes achieves both human indecipherability and machine decipherability excellently. The source codes are available at \url{https://github.com/mcljtkim/Forbes}.
\end{abstract}

\section{Introduction}
\label{sec:intro}

\begin{figure}[t!]
\begin{minipage}[h]{0.48\linewidth}
\centering
\includegraphics[width=5.9cm,height=2.6cm]{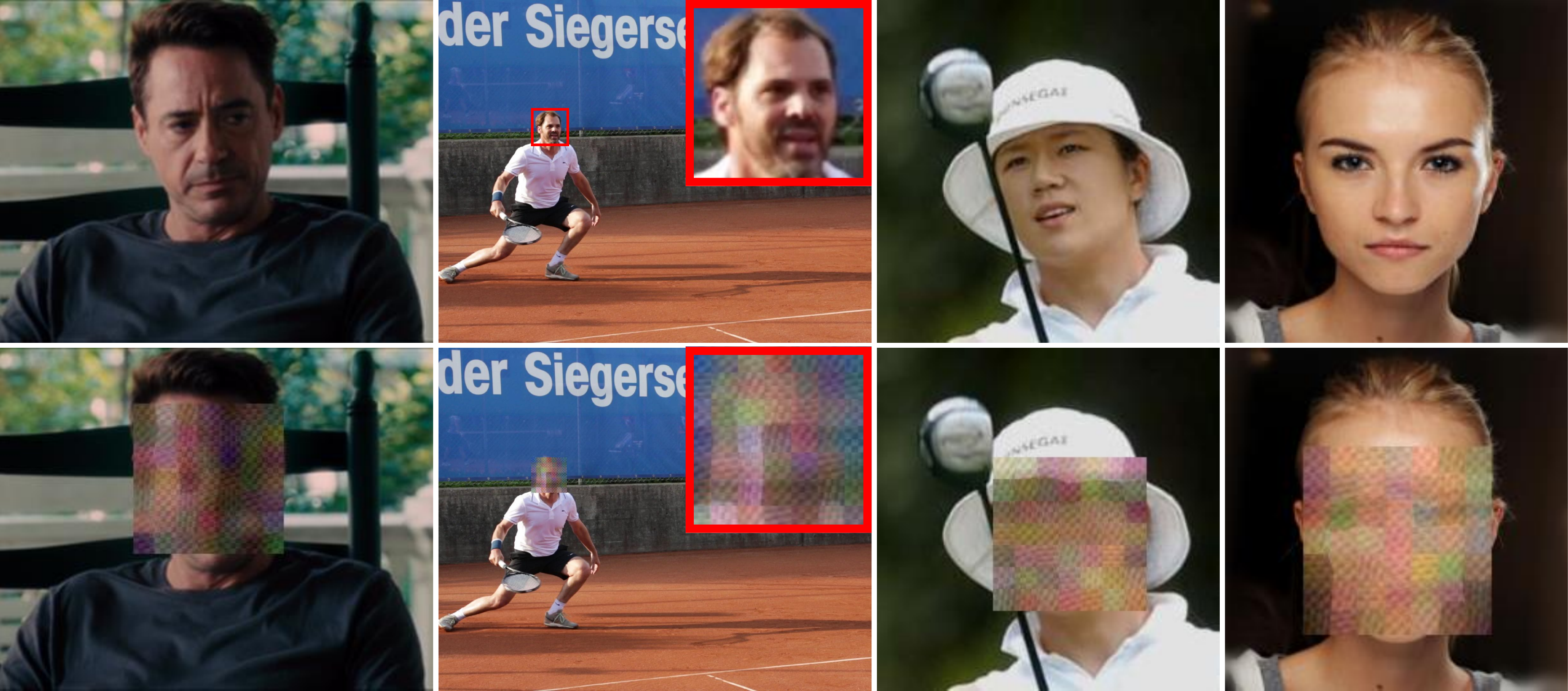}
\captionof{figure}{Top, facial images are selected from the Vimeo-90K \cite{xue2019toflow}, DAVIS 2016 \cite{Perazzi2016davis}, LFW \cite{huang2008lfw}, CelebA \cite{liu2015celeba} datasets. Bottom, the facial regions are obfuscated by the proposed Forbes algorithm.}
\label{fig:intro}
\end{minipage}\hfill
\begin{minipage}[h]{0.48\linewidth}
\centering
\includegraphics[width=5.9cm,height=2.6cm]{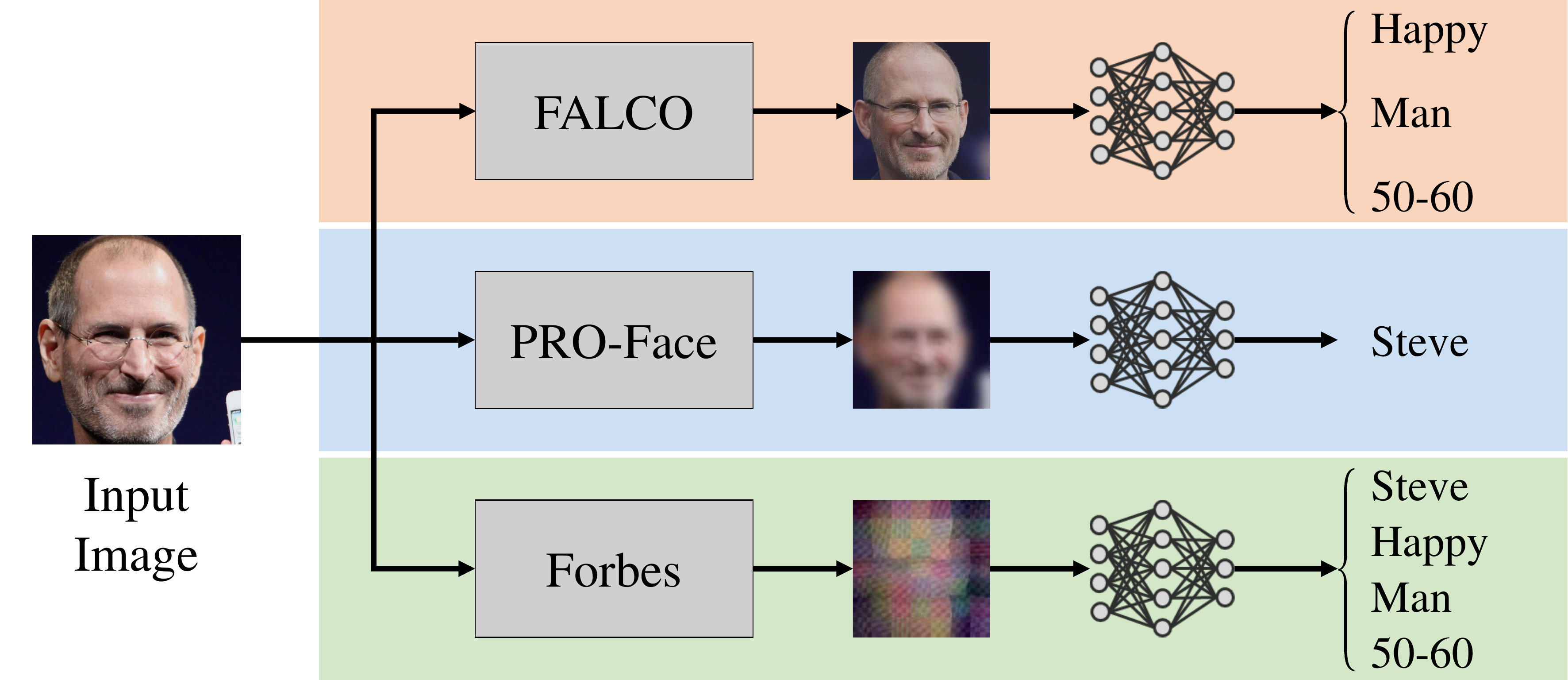}
\captionof{figure}{Illustration of differences between the existing attribute-preserving face anonymization algorithm \cite{barattin2023attribute}, the existing face obfuscation algorithm \cite{yuan2022pro}, and the proposed Forbes algorithm.}
\label{fig:related}
\end{minipage}
\end{figure}

Deep learning networks are widely used for various vision tasks \cite{sun2018pwc, jun2022depth, dong2015srcnn}. To train such networks effectively, a huge amount of high-quality data is required in general. Hence, various datasets have been proposed, but the privacy of individuals included in the datasets should be protected \cite{voigt2017eu}. Fig.~\ref{fig:intro} shows some images in popular datasets \cite{xue2019toflow, Perazzi2016davis, CelebAMask-HQ, huang2008lfw}, containing faces. Although privacy issues can be raised for any images, they occur mostly for facial images. As illustrated in the bottom row of Fig.~\ref{fig:intro}, we obfuscate facial regions for privacy protection.

Facial image datasets are used for diverse tasks, such as face verification \cite{deng2019arcface, kim2022adaface, meng2021magface}, gender classification \cite{hung2019compacting}, race classification \cite{sarridis2023flac}, age estimation \cite{shin2022mwr, lee2022gol, lee2020deep}, and facial expression recognition \cite{zhao2021robust, lee2024unsupervised, lee2022order}. Recently, to free such datasets from privacy concerns, privacy-enhancing technologies have been developed. For example, there are face anonymization methods \cite{Rosberg2023fiva, ciftci2023mfmc, barattin2023attribute, sun2018natural, hukkelaas2019deepprivacy} based on generative models, which replace original faces with fake ones to make the individuals' identities unrecognizable by both humans and machines. In contrast to face anonymization, face obfuscation \cite{li2021learning, yuan2022pro, li2021identity, li2023privacy} attempts to retain the identities of individuals for machines, as illustrated in Fig.~\ref{fig:related}. An obfuscated image should be decipherable by deep learning networks but unrecognizable by humans. In other words, an obfuscated image should satisfy two properties: human indecipherability (HI) and machine decipherability (MD).

In this paper, we propose a novel face obfuscation algorithm, called \textbf{F}ace \textbf{o}bfuscation \textbf{r}endering via \textbf{b}ack\-prop\-agation r\textbf{e}finement \textbf{s}cheme (Forbes), which achieves both HI and MD. As in Fig.~\ref{fig:related}, whereas the existing obfuscation methods \cite{li2021learning, yuan2022pro, li2021identity, li2023privacy} retain only identity information for machines, Forbes preserves attributes as well. We first develop eight parameterized obfuscating transformations. Then, we apply those transformations with the parameters to an input image to remove the identity information recognizable by humans. Then, we refine the parameters to make the transformed image decipherable by machines. Specifically, to balance HI and MD, we define energy functions for face obfuscation, minimize them using the backpropagation refinement scheme (BRS) \cite{jang2019interactive}, and obtain optimal parameters. We then generate an obfuscated image by applying the transformations with the optimal parameters. Extensive experiments demonstrate that the proposed Forbes achieves both HI and MD excellently.

This work has the following major contributions:
\begin{itemize}
  \item To the best of our knowledge, Forbes is the first attempt to encrypt both identity and facial attribute information, decipherable by machines but unrecognizable by humans.
  \item We introduce two HI energy functions and two MD energy functions and minimize them based on BRS to strike a tradeoff between HI and MD.
  \item Forbes provides excellent obfuscating results on various benchmark datasets for face verification, age estimation, gender classification, and facial attribute classification.
\end{itemize}

\section{Related Work}
\label{sec:related}

\subsection{Face Anonymization}

Face anonymization is a vision task to remove the identity of a person from the image. It can be achieved by simple filters, such as blurring, masking, or mosaicking. Early anonymization methods \cite{hasan2017cartooning, brkic2017know} are based on these filters. Specifically, Hassan \etal \cite{hasan2017cartooning} masked privacy-sensitive regions, including faces, with clip art. Also, Brki\'{c} \etal \cite{brkic2017know} blurred a person's face for identity removal but maintained the person's pose and clothes. Filtering methods, however, have a limitation in that their anonymized images may be restored by inverse filters~\cite{li2023simple, yasarla2020deblurring}.

Recent anonymization methods \cite{cho2020cleanir, chen2018vgan,ren2018learning, sun2018natural, hukkelaas2019deepprivacy, maximov2020ciagan, ciftci2023mfmc, Rosberg2023fiva, barattin2023attribute} have adopted generative models to synthesize identity-removed faces.
For instance, Ren \etal \cite{ren2018learning} proposed a video anonymizer, which removes the identities of people while preserving their actions. Also, Chen \etal \cite{chen2018vgan} incorporated a generative model into an auto-encoder for anonymization, which preserves facial expressions (\eg happiness and surprise). Cho \etal \cite{cho2020cleanir} developed an auto-encoder to decompose feature vectors into identity features and attribute features (\eg gender and race). They modified the identity features but left the attribute features unaltered. Thus, their method can perform attribute-preserving face anonymization. Barattin \etal \cite{barattin2023attribute} proposed another attribute-preserving anonymization method via latent code optimization. Since facial attributes are useful in various applications, such as expression recognition \cite{zhao2021robust} and age estimation \cite{lee2022gol, shin2022mwr}, we also attempt to preserve facial attributes in the proposed obfuscation algorithm.

\subsection{Face Obfuscation}
Face obfuscation encrypts the identity of a person in an image so that only a machine can decrypt the identity (\ie MD) while humans cannot recognize it (\ie HI). Note that obfuscation is different from anonymization in that the former aims to retain some identity information decipherable by machines only, whereas the latter tries to remove all identity information and make it indecipherable by both humans and machines, as in Fig.~\ref{fig:related}. Thus, face obfuscation is essential in some applications, such as re-identification with privacy protection. However, relatively few algorithms have been developed for face obfuscation.

In an early study, Vishwamitra \etal \cite{vishwamitra2017blur} focused on HI only and suggested that masking a person with a black box is more effective than blurring. Li \etal \cite{li2021learning} first adopted an auto-encoder for face obfuscation. They separated identity features from attribute ones, as in \cite{cho2020cleanir}, and then performed obfuscation by replacing the attributes with different person's attributes. Li \etal \cite{li2021identity} also proposed adaptive attribute obfuscation method. They extracted facial attributes, identified important attributes for MD, and modified less important ones. However, since they focused on MD only, their obfuscated images are often recognizable by humans as well. To improve HI, Yuan \etal \cite{yuan2022pro} proposed a deep fusion network. They first pre-obfuscated an image using pre-defined operations, such as blurring or pixelating, making it difficult for humans to recognize the identity. Then, they fused the pre-obfuscated image with the original image to yield a final obfuscation result. Although their algorithm achieves a high accuracy when tested by a face verification network, their obfuscated image still retains facial appearance decipherable by humans, as shown in Fig.~\ref{fig:related}. In contrast, the proposed Forbes algorithm makes the resultant image much less decipherable by humans. Furthermore, Forbes is attribute-preserving; from its obfuscated image, a machine can decipher facial attributes as well as identity.

\begin{figure}[t]
    \centering
    \includegraphics[width=\linewidth]{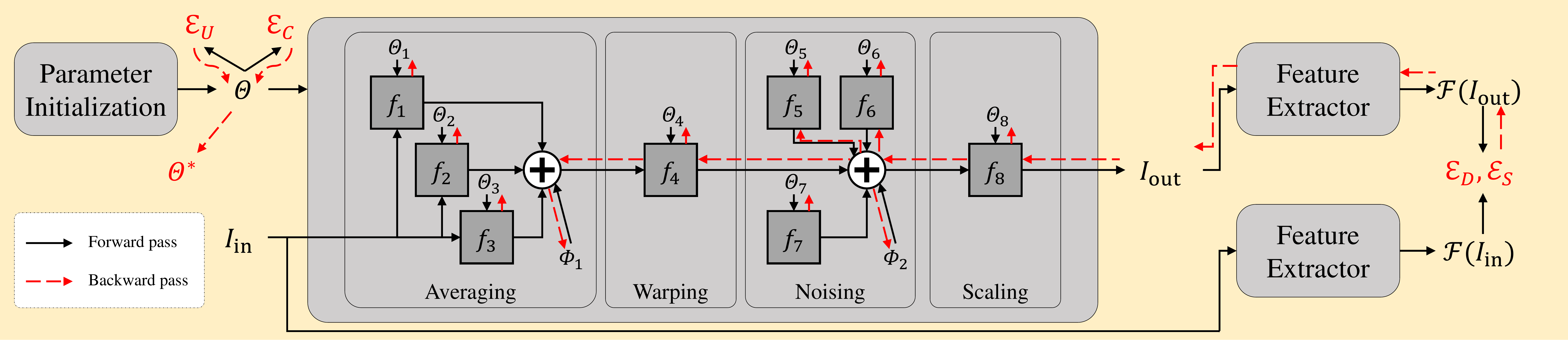}
    \caption{An overview of the proposed Forbes algorithm. Each obfuscating transformation $f_n$ has a set $\Theta_n$ of parameters. Also, $\Phi_1$ and $\Phi_2$ are composed of superposing parameters. A randomly initialized set $\Theta = \Theta_1 \cup \cdots \cup \Theta_8 \cup \Phi_{\rm 1} \cup \Phi_{\rm 2}$ is optimized to $\Theta^*$ in the backward pass to balance HI and MD. To this end, energy functions $\mathcal{E}_U$, $\mathcal{E}_C$, $\mathcal{E}_D$, and  $\mathcal{E}_S$ are minimized.}
    \label{fig:overview}
\end{figure}

\section{Proposed Algorithm}
\label{sec:proposed}
Forbes uses several transformations to obfuscate an image. Each transformation has parameters to control the balance between HI and MD. We apply these transformations to process an image, after initializing their parameters randomly within their ranges. The transformations are designed, in general, to make an output image not easily recognizable by humans. However, the output may be also indecipherable by machines, such as face verification networks \cite{kim2022adaface, meng2021magface}. To improve the MD, we optimize the transformation parameters to minimize the distance and maximize the cosine similarity between the two feature vectors extracted from the input and output images, respectively, by a feature extractor. For the optimization, we adopt the BRS technique \cite{jang2019interactive}. Note that the parameters of the feature extractor are fixed during the optimization process. Finally, we render an obfuscated image using the transformations with the optimal parameters.

Fig.~\ref{fig:overview} is an overview of the Forbes algorithm. Let us describe each algorithm component subsequently.

\subsection{Obfuscating Transformations}
\label{ssec:tran}

\begin{figure}[t]
    \centering
    \includegraphics[width=\linewidth]{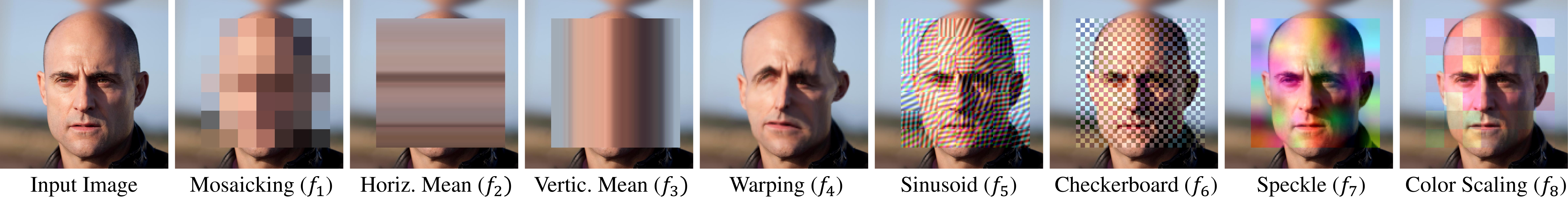}
    \caption{Examples of transformed images by $f_n$.}
    \label{fig:trans_example}
\end{figure}

We design eight face obfuscating transformations $f_n$, which are listed in Table~\ref{table:transformation}, in four categories: 1) averaging, 2) warping, 3) noising, and 4) scaling. Each transformation $f_n$ has parameters. Let $\Theta_n$ denote the set of parameters for $f_n$. Fig.~\ref{fig:trans_example} shows examples of transformed images.

\textbf{Averaging:} In the averaging category, we process an input RGB image $I_{\rm in} \in[0, 1]^{H\times W\times 3}$ to generate a degraded image $I_{\rm deg} \in[0, 1]^{H\times W\times 3}$ that retains low-frequency information of the input image while removing its high-frequency information. For example, in \textit{Mosaicking} ($f_1$), we divide $I_{\rm in}$ into $M\times N$ blocks of resolution $\frac{H}{M} \times \frac{W}{N}$ and fill in each block with its average value to yield $I_{\rm deg}$. Similarly, \textit{Horizontal Mean} ($f_2$) and \textit{Vertical Mean} ($f_3$) fill in each pixel with the average value of the $W$ horizontal pixels and the $H$ vertical pixels, respectively. Let $B_{\rm in}$ and $B_{\rm deg}$ denote a block in $I_{\rm in}$ and its corresponding block in $I_{\rm deg}$, respectively. Then, we obtain an output block by
\begin{equation}
    B_{\rm out}^c = (1-\theta^c)\cdot B_{\rm in}^c + \theta^c \cdot B_{\rm deg}^c
    \label{eq:ble}
\end{equation}
where $c$ denotes one of the three RGB channels, and $\theta^c$ is a blending parameter. To remove the high-frequency information effectively, we fix the blending parameters to $1$, as specified in Table~\ref{table:transformation}.

\begin{table*}[t]
    \caption{Specification of the eight obfuscating transformations: $\theta$-range denotes the range of each parameter $\theta$ in $\Theta_n$, and $|\Theta_n|$ is the number of parameters in $\Theta_n$. Also, $\lambda$ is a margin for the HI energy.}
    \centering
    \resizebox{0.95\linewidth}{!}{
    \begin{tabular}{L{3.0cm}L{0.6cm}L{3.0cm}C{2.4cm}C{4.0cm}C{1.2cm}} \toprule
    Category                   & \multicolumn{2}{C{1.65cm}}{Transformation}  & $\theta$-range  & $|\Theta_n|$  & $\lambda$ \\ \midrule
    \multirow{3}{*}{Averaging} & $f_{1}$ & Mosaicking     & $[1, 1]$         & $M\times N\times 3$        & $0$       \\
                               & $f_{2}$ & Horizontal Mean& $[1, 1]$         & $M\times N\times 3$        & $0$       \\
                               & $f_{3}$ & Vertical Mean  & $[1, 1]$         & $M\times N\times 3$        & $0$       \\ \midrule
    Warping                    & $f_{4}$ & Warping        & $[-0.3, 0.3]$    & $(M-1)\times (N-1)\times2$ & $0.05$    \\ \midrule
    \multirow{3}{*}{Noising}   & $f_{5}$ & Sinusoid       & $[0, \pi]$       & $M\times N\times 3$        & $0$       \\
                               & $f_{6}$ & Checkerboard   & $[1, 1]$         & $M\times N\times 3$        & $0$       \\
                               & $f_{7}$ & Speckle        & $[-0.5, 0.5]$    & $M\times N\times 3$        & $0.1$     \\ \midrule
    Scaling                    & $f_{8}$ & Color Scaling  & $[10/11, 11/10]$ & $M\times N\times 3$        & $1.05$    \\ \bottomrule
    \end{tabular}
    }
    \label{table:transformation}
\end{table*}

\textbf{Warping:} To modify $I_{\rm in}$ geometrically, we adopt a geometric transformation, \textit{Warping} ($f_4$). We use the $(M-1)\times (N-1)$ inner corners of the $M\times N$ blocks as grid points and perform grid warping to distort $I_{\rm in}$. Each grid point is moved by $(\theta_x \Delta_x, \theta_y  \Delta_y)$, where $\Delta_x$ and $\Delta_y$ are the horizontal and vertical sizes of the block. Both warping parameters $\theta_x$ and $\theta_y$ are selected from $[-0.3, 0.3]$ to prevent grid folding.

\textbf{Noising:} In the noising category, we obtain an output image by
\begin{equation}
    I_{\rm out} = I_{\rm in} + \mathcal{N}
\end{equation}
where $\mathcal{N}$ is a noise map. Similar to $f_1$, we first divide $I_{\rm in}$ into $M\times N$ blocks. Then, in \textit{Sinusoid} ($f_5$), for each block, we generate a sinusoidal noise along an axis at a random angle $\theta^c$. The period of the sinusoid is a quarter of $\min \{ \Delta_x, \Delta_y \}$, and the amplitude is 0.2. In \textit{Checkerboard} ($f_6$), for each block, a $4\times 4$ checkerboard pattern is used as noise, as shown in Figure~\ref{fig:trans_example}. The absolute magnitude of the pattern is $0.3 \times \theta^c$ for each color channel, and the mean value is 0. In \textit{Speckle} ($f_7$), we assign parameter $\theta^c$ to the center pixel of each block, and then bilinearly interpolate the remaining pixels from the center ones. Thus, the high-frequency components in an output image $I_{\rm out}$ may be strengthened, for the noise map $\mathcal{N}$ is composed of fluctuating patterns.

\textbf{Scaling:} For obfuscation, we also modulate colors with \textit{Color Scaling} ($f_8$). Specifically, we obtain an output block by
\begin{equation}\label{eq:color}
B_{\rm out}^c = {\theta^c} \cdot B_{\rm in}^c
\end{equation}
where $\theta^c$ is a scaling parameter. Since $f_8$ just modifies the scales of the color components, high-frequency information in an input image $I_{\rm in}$ can be preserved in the output image $I_{\rm out}$.

\subsection{Composite Transformation}
\label{ssec:composite}

We combine the eight obfuscating transformations $f_n$, $1\leq n \leq 8$, into a single composite transformation. Note that, when only one obfuscating transformation is used as in Fig.~\ref{fig:trans_example}, the output image may contain some identity information still recognizable by humans, or erase most of the information decipherable by machines. To overcome this issue, we superpose the output images of the transformations in the same category.

First, we apply $f_1$, $f_2$, and $f_3$ to the input image and superpose the three output images in a blockwise manner. Specifically, let $B_1$, $B_2$, and $B_3$ denote a triplet of corresponding blocks in these output images. We normalize random parameters $\phi_1^c$, $\phi_2^c$, $\phi_3^c \in \Phi_1$ into $\bar{\phi}_1^c$, $\bar{\phi}_2^c$, and $\bar{\phi}_3^c$ using the softmax operator and obtain the convex combination
\begin{equation}
  B_{\rm avg}^c = \bar{\phi}_1^c \cdot B_1^c + \bar{\phi}_2^c \cdot B_2^c + \bar{\phi}_3^c \cdot B_3^c
  \label{eq:convex}
\end{equation}
where $c$ is the RGB channel index. $I_{\rm avg}$ comprises these superposed blocks. Then, we apply $f_4$ to $I_{\rm avg}$ to yield a warped image $I_{\rm warped}$. Since $I_{\rm warped}$ contains only low-frequency information, we generate high-frequency information with the noising transformations to enable a machine to decipher the identity. Similarly to \eqref{eq:convex}, we normalize random parameters $\phi_5^c$, $\phi_6^c$, $\phi_7^c \in \Phi_2$ into $\bar{\phi}_5^c$, $\bar{\phi}_6^c$, and $\bar{\phi}_7^c$, and superpose the noise blocks with the warped block by
\begin{equation}
  B_{\rm noi}^c = B_{\rm warped} + \bar{\phi}_5^c \cdot B_5^c + \bar{\phi}_6^c \cdot B_6^c + \bar{\phi}_7^c \cdot B_7^c
  \label{eq:convex2}
\end{equation}
where $B_n^c$ is a noise block from $f_n$. $I_{\rm noi}$ is composed of these superposed blocks. Last, we apply the scaling transformation $f_8$ to $I_{\rm noi}$ to yield the final output image $I_{\rm out}$.

\subsection{Parameter Initialization}
\label{ssec:init}

Let $\Theta$ denote the set of all parameters for the composite transformation. It consists of four kinds of parameters;
\begin{equation}
\Theta = \Theta_F \cup \Theta_U \cup \Theta_C \cup \Phi
\end{equation}
where  $\Theta_F$ is the set of fixed parameters, $\Theta_U$ is the set of uniform parameters, $\Theta_C$ is the set of color parameters, and $\Phi$ is the set of composing parameters. Note that $\Theta_F=\Theta_1 \cup \Theta_2 \cup \Theta_3 \cup \Theta_6$, $\Theta_U=\Theta_4 \cup \Theta_5 \cup \Theta_7$, $\Theta_C=\Theta_8$, and $\Phi=\Phi_1 \cup \Phi_2$.

\textbf{Fixed parameters}: The parameters of the four transformations $f_1$, $f_2$, $f_3$, and $f_6$ are fixed to $1$. For $f_1$, $f_2$, or $f_3$, $\theta^c$ is fixed to 1 to erase the high-frequency information or image details with a strong low-pass filter. In contrast, for $f_7$, the amplitude of the checkerboard pattern is fixed to $0.3\times\theta^c = 0.3$ experimentally to mimic the high-frequency information in the original image and achieve MD.

\textbf{Uniform parameters}: For the three transformations $f_4$, $f_5$, and $f_7$, the parameters are initialized uniformly. Note that, in Table~\ref{table:transformation}, $\theta$-range specifies the range $[\theta_{\rm min}, \theta_{\rm max}]$ of each parameter $\theta$ in $f_n$. We initialize $\theta$ according to the uniform distribution ${\cal U}(\theta_{\rm min}, \theta_{\rm max})$ with the probability density function, given by
\begin{equation}
  p(\theta)=
  \begin{cases}
  \frac{1}{\theta_{\rm max}-\theta_{\rm min}} & \mbox{if } \theta_{\rm min} \le \theta \le \theta_{\rm max},\\
  0 & \mbox{otherwise}.
  \end{cases}
\end{equation}

\textbf{Color parameters}: For \textit{Color Scaling} ($f_8$), an input color is multiplied by a parameter $\theta$ to yield an output color in \eqref{eq:color}. If $\theta>1$, it is a brightening function. On the contrary, if $0<\theta<1$, it is a darkening one. Hence, we first decide with the equal probabilities of $\frac{1}{2}$ whether to choose a brightening or darkening function. Then, for a brightening one, the scaling parameter $\theta$ is initialized according to the uniform distribution ${\cal U}(1, \theta_{\rm max})$. On the other hand, for a darkening one, the inverse $\frac{1}{\theta}$ is initialized according to ${\cal U}(1, \theta_{\rm max})$. Thus, $\theta$ is determined within $[\frac{1}{\theta_{\rm max}}, \theta_{\rm max}]$. As specified in Table~\ref{table:transformation}, $\theta_{\rm max}=\frac{11}{10}$.

\textbf{Composing parameters}: Each composing parameter $\phi$ for superposing the results of $f_1$, $f_2$, and $f_3$ in \eqref{eq:convex} or $f_5$, $f_6$, and $f_7$ in \eqref{eq:convex2} is initialized according to the uniform distribution ${\cal U}(0, 1)$. Note that these composing parameters are normalized to $\bar{\phi}$ using the softmax operator for the convex combination.

\begin{figure}[t]
    \centering
    \includegraphics[width=\linewidth]{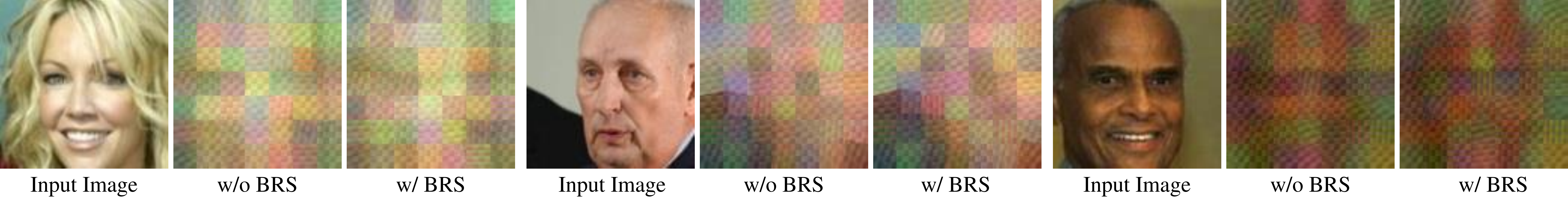}
    \caption{Comparison of obfuscated images with or without the BRS optimization.}
    \label{fig:optimization}
\end{figure}

\subsection{Parameter Optimization}
\label{ssec:opt}

A transformed image $I_{\rm out}(\Theta)$, which is obtained by applying the composite transformation with randomly initialized parameters in $\Theta$ to an input image $I_{\rm in}$, may be indecipherable by a machine. More specifically, when it is processed by a feature extractor, its output vector $\mathcal{F}(I_{\rm out}(\Theta))$ may deviate from the original vector $\mathcal{F}(I_{\rm in})$ excessively, yielding a wrong decision result, \eg a wrong identity. Hence, to improve the MD, we adopt the BRS technique \cite{jang2019interactive} to optimize the parameters in $\Theta$ so that the distance $\| \mathcal{F}(I_{\rm out}(\Theta)) - \mathcal{F}(I_{\rm in})\|_2$ between the two feature vectors is minimized and their cosine similarity $\frac{\mathcal{F}(I_{\rm out}(\Theta))^T\mathcal{F}(I_{\rm in})}{\| \mathcal{F}(I_{\rm out}(\Theta))\| \|\mathcal{F}(I_{\rm in})\|}$ is maximized. Fig.~\ref{fig:optimization} shows examples of the BRS optimization. We see that the optimization does not change the level of HI noticeably. However, it improves the degree of MD effectively, as will be shown in Section \ref{sec:analysis}.

BRS \cite{jang2019interactive} was originally proposed for interactive image segmentation to optimize an interaction map to refine a segmentation result. In this work, we adopt BRS to optimize the parameters in $\Theta$ of the composite transformation. To this end, we define an energy function $\mathcal{E}(\Theta)$ appropriate for face obfuscation and develop its minimization scheme.

For uniform parameters, we have two types of $\theta$-ranges in Table~\ref{table:transformation}: $[\theta_{\rm min}, \theta_{\rm max}]$ or $[-\theta_{\rm max}, \theta_{\rm max}]$. In both types, the obfuscating functions are designed so that HI increases, or at least maintains the same level, as $|\theta|$ gets closer to $\theta_{\rm max}$. Thus, to improve HI, we define the HI energy for uniform parameters as
\begin{equation}\label{eq:margin_u}
  \mathcal{E}_{U} = \sum_{\theta_i \in \Theta_i \subset \Theta_U}{\rm max} \{\lambda_i - |\theta_i|, 0\}
\end{equation}
where $\lambda_i$ is the margin for $\Theta_i$ specified in Table \ref{table:transformation}. In BRS, the partial derivative of $\mathcal{E}_U$ with respect to each uniform parameter $\theta_i$ is required, which is given by
\begin{equation}\label{eq:pd_margin_u}
    {\partial \mathcal{E}_U \over \partial \theta_i}=
    \begin{cases}
        -1 & \mbox{if } 0<\theta_i<\lambda_i, \\
        1 &  \mbox{if } -\lambda_i<\theta_i<0, \\
        0 &  \mbox{otherwise}.
    \end{cases}
\end{equation}
Similarly, we define the HI energy for color parameters as
\begin{equation}\label{eq:margin_p}
  \mathcal{E}_C = \sum_{\theta_i \in \Theta_i \subset \Theta_C}{\rm max}\big\{[\theta_i>1](\lambda_i - \theta_i), [\theta_i<1](\theta_i -  \frac{1}{\lambda_i})\big\}
\end{equation}
where $[\cdot]$ is an indicator function. Note that, if $\theta_i=1$, the corresponding color is not scaled in $f_8$. Hence, by minimizing $\mathcal{E}_C$, it is encouraged that $\theta_i$ is as different from $1$ as possible. The derivative of $\mathcal{E}_C$ is given by
\begin{equation}\label{eq:pd_margin_p}
    {\partial \mathcal{E}_C \over \partial \theta_i}=
    \begin{cases}
        -1 & \mbox{if } 1<\theta_i<\lambda_i,\\
        1 & \mbox{if } {1\over \lambda_i}<\theta_i<1,\\
        0 & \mbox{otherwise}.
    \end{cases}
\end{equation}

On the other hand, to improve MD, the distance between the feature vectors should be reduced. Hence, we define the MD energy of the distance $\mathcal{E}_D$ and the MD energy of the cosine similarity $\mathcal{E}_S$ as
\begin{eqnarray}
   \mathcal{E}_D&=&\|\mathcal{F}(I_{\rm out}(\Theta)) - \mathcal{F}(I_{\rm in})\|_2, \label{eq:md_energy1} \\
   \mathcal{E}_S&=&1-\frac{\mathcal{F}(I_{\rm out}(\Theta))^T\mathcal{F}(I_{\rm in})}{\| \mathcal{F}(I_{\rm out}(\Theta))\| \|\mathcal{F}(I_{\rm in})\|}. \label{eq:md_energy2}
\end{eqnarray}

Then, we adopt the overall energy function given by
\begin{equation}\label{eq:energy}
        \mathcal{E}=\mathcal{E}_U + \mathcal{E}_C + \mathcal{E}_D + \mathcal{E}_S
\end{equation}
whose derivative is
\begin{equation}\label{eq:pd_energy}
    {\partial\mathcal{E}\over \partial\Theta} = {\partial\mathcal{E}_U \over\partial\Theta}+{\partial\mathcal{E}_C \over\partial\Theta}+{\partial\mathcal{E}_D \over\partial\Theta} + {\partial\mathcal{E}_S \over\partial\Theta}.
\end{equation}
The first and second terms can be computed using \eqref{eq:pd_margin_u} and \eqref{eq:pd_margin_p}, respectively, while the last two terms ${\partial\mathcal{E}_D \over\partial\Theta}$ and ${\partial\mathcal{E}_S \over\partial\Theta}$ are computed by BRS.
Thus, we minimize $\mathcal E$ using the L-BFGS method \cite{liu1989limited} and determine an optimal set of parameters $\Theta^*$. However, the fixed parameters are not modified in the optimization step, so $\Theta^*_F=\Theta_F$.

\begin{table*}[t]
    \caption{Quantitative comparison of face obfuscation results. In each test, the best result is \textbf{boldfaced}, while the second best is \underline{underlined}.}
    \renewcommand{\arraystretch}{1.1}
    \centering
    \resizebox{1.0\linewidth}{!}
    {\begin{tabular}{L{1.0cm}R{3.0cm}C{1.4cm}C{2.0cm}C{2.0cm}C{2.3cm}C{2.6cm}C{2.8cm}}\toprule
         & & Human & \multicolumn{2}{C{4.0cm}}{Face Verification} & Age & Gender & Attribute \\ \cmidrule(lr){3-3} \cmidrule(lr){4-5} \cmidrule(lr){6-6} \cmidrule(lr){7-7} \cmidrule(lr){8-8}
         & Dataset $\vartriangleright$& LFW & LFW & IJB-C & Adience & Adience & CelebAMask-HQ  \\ \cmidrule(lr){3-8}
         & Metric $\vartriangleright$                                     & ACR ($\downarrow$) & Acc. ($\uparrow$)& Acc. ($\uparrow$)& Acc. ($\uparrow$)& Acc. ($\uparrow$)& Acc. ($\uparrow$)\\ \midrule
         \multicolumn{2}{L{4.0cm}}{Original}                              &  --   &99.82&94.64&65.24&85.65&88.02\\ \midrule
         \multicolumn{2}{L{4.0cm}}{Blur}                                  &0.8378&79.20&81.82&43.33&31.36&61.31\\
         \multicolumn{2}{L{4.0cm}}{Pixelate}                              &0.7622&85.58&83.54&22.60&64.97&62.74\\
         \multicolumn{2}{L{4.0cm}}{FALCO \cite{barattin2023attribute}}    &0.6844&78.68&71.51&20.24&71.53&72.88\\
         \multicolumn{2}{L{4.0cm}}{PRO-Face (Blur) \cite{yuan2022pro}}    &0.8978&\underline{98.10}&87.63&\underline{57.01}&40.86&65.04\\
         \multicolumn{2}{L{4.0cm}}{PRO-Face (Pixelate) \cite{yuan2022pro}}&0.8133&94.35&86.93&23.66&72.18&66.79\\
         \multicolumn{2}{L{4.0cm}}{PRO-Face (FaceShifter) \cite{yuan2022pro}}&0.8333&96.72&80.23&21.14&75.68&75.13\\
         \multicolumn{2}{L{4.0cm}}{PRO-Face (SimSwap) \cite{yuan2022pro}} &0.7800&96.13&78.43&21.51&65.43&75.47\\ \midrule
         \multicolumn{2}{L{4.0cm}}{Forbes-G}                              &\underline{0.2578}&96.10&\underline{89.47}&49.82&\underline{78.64}&\underline{73.18}\\
         \multicolumn{2}{L{4.0cm}}{Forbes-T}                              &\textbf{0.2422}&\textbf{98.72}&\textbf{90.89}&\textbf{59.03}&\textbf{81.26}&\textbf{76.20}\\ \bottomrule
    \end{tabular}
    }
    \label{tab:quantitative}
\end{table*}

\subsection{Implementation Details}

It is worth pointing out that the proposed Forbes algorithm operates at the inference stage, so it does not need any training. Forbes divides a facial region into $7\times 7$ blocks. In other words, we set $M=N=7$. We analyze the impacts of $M$ and $N$ on the obfuscating performance in Section \ref{sec:analysis}.

Forbes aims to generate an obfuscated image that can be deciphered by machines only, so we may use several pre-trained networks in parallel as the feature extractor in Fig.~\ref{fig:overview}. However, the inference time of Forbes depends on the feature extractor, since Forbes should compute the derivative of the energy function. If we combine more pre-trained networks in parallel, the obfuscated image may retain more information decipherable by machines, but the parameter optimization becomes slower in general. Therefore, to reduce the inference time, we may use a task-specific feature extractor, instead of several networks. Thus, we implement two types of Forbes: Forbes-G and Forbes-T.

\subsubsection{Forbes-G:} In Forbes-G, we employ five pre-trained networks in parallel as the feature extractor regardless of the task. Specifically, we use pre-trained AdaFace~\cite{kim2022adaface}, GOL~\cite{lee2022gol}, SphereNet~\cite{hung2019compacting}, MobileNetV2~\cite{sandler2018mobilenetv2}, and ResNet-50~\cite{he2016deep} as the feature extractor. Thus, the output image generated by Forbes-G tends to contain identity, age, gender, and facial attributes.

\subsubsection{Forbes-T:} Depending on the required information for a given task, we employ a task-specific feature extractor in Forbes-T. For example, we adopt the feature extractor of a face verification network to retain the identity information or that of a gender classification network to retain the gender information.

\section{Experiments}
We assess Forbes on various benchmark datasets for face-related vision tasks.

\begin{figure*}[t]
    \centering
    \includegraphics[width=\linewidth]{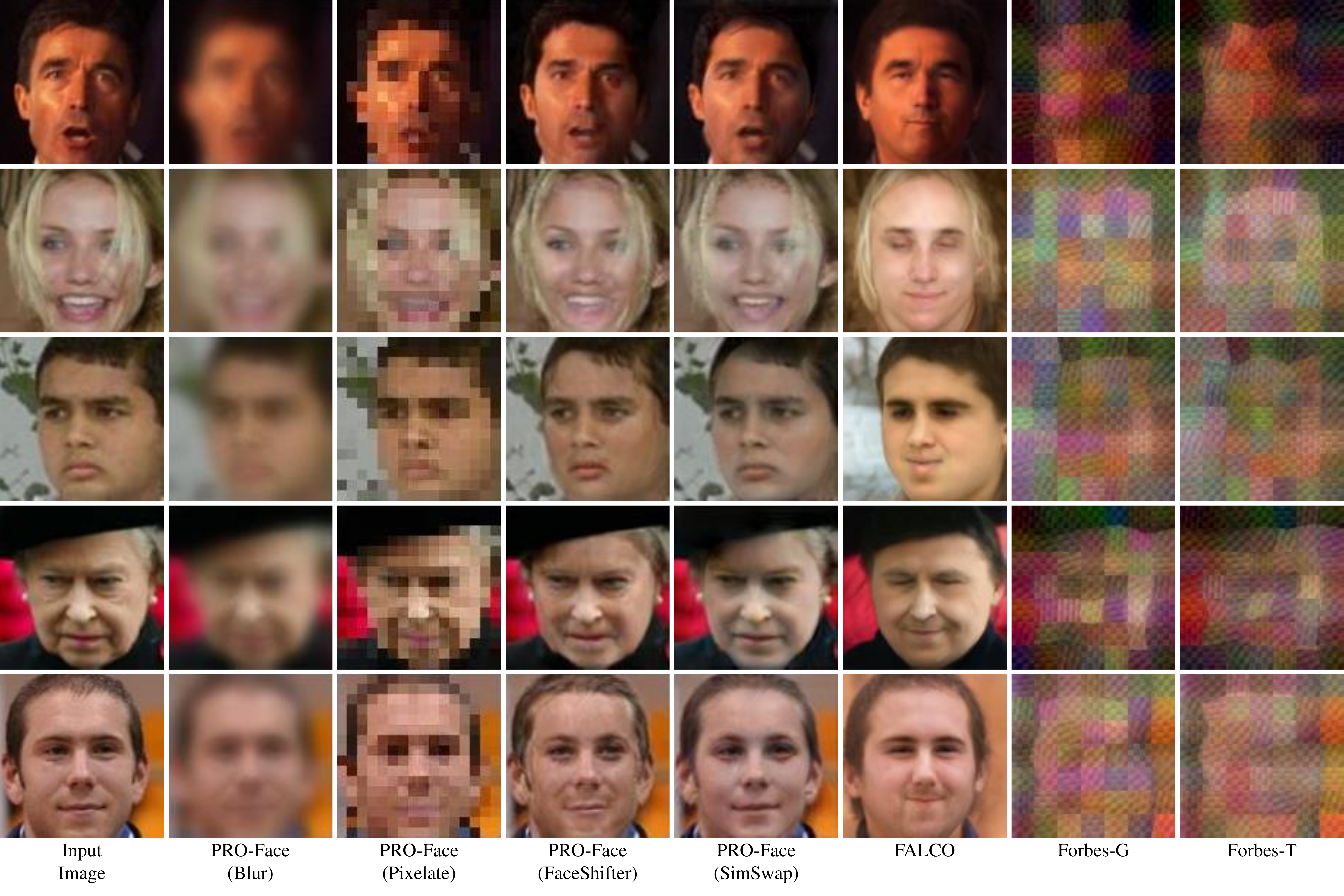}
    \caption{Qualitative comparison of obfuscated images in the LFW dataset.}
    \label{fig:qual_fv}
\end{figure*}

\subsection{Datasets and Evaluation Protocols}
\textbf{Face verification:} LFW \cite{huang2008lfw} consists of 13,233 facial images with 5,749 identities. We use the test set containing 6,000 pairs of facial images with half matched and half unmatched. IJB-C \cite{maze2018ijbc} consists of 138,000 facial images. We randomly sample 20,000 pairs with half matched and half unmatched. To use the existing face verification network \cite{kim2022adaface}, we align $112\times112$ facial regions using landmarks detected by MTCNN \cite{zhang2016mtcnn}. We measure the verification accuracy of AdaFace \cite{kim2022adaface} using their official model. We verify whether two facial images, reference $I_{\rm ref}$ and source $I_{\rm out}$, belong to the same identity. Note that $I_{\rm out}$ is obfuscated from an input image $I_{\rm in}$ that is different from $I_{\rm ref}$. This face verification tests demonstrate the identity-preserving capability of Forbes.

\textbf{Age estimation \& gender classification}: For these two tasks, we use the Adience dataset \cite{eidinger2014adience}, composed of 26,580 facial images of 2,284 identities. For age estimation, we measure the accuracy score using the GOL algorithm \cite{lee2022gol}, which performs the k-NN estimation in a feature space. We train GOL with the provided settings and conduct the 5-fold cross-validation on Adience \cite{eidinger2014adience, shin2022mwr}. We use the official model of SphereNet \cite{hung2019compacting} for gender classification.

\textbf{Facial attribute classification:} CelebAMask-HQ \cite{CelebAMask-HQ} contains 30,000 cele\-brity images selected from CelebA \cite{liu2015celeba}. Each face of resolution $1024\times1024$ is annotated with 40 attribute labels (\eg eyeglasses, hair color, and mustache). We test the capability of retaining facial attributes after obfuscation. For attribute classification, we train MobileNetV2 \cite{sandler2018mobilenetv2} on CelebA.

\subsection{Comparative Results}

Table~\ref{tab:quantitative} compares the accuracies of Forbes on face verification, age estimation, gender classification, and facial attribute classification with those of previous methods \cite{barattin2023attribute, yuan2022pro}. Note that Forbes-T significantly outperforms the existing algorithms on all the tasks. In particular, compared to the original images, Forbes-T degrades the face verification accuracies only marginally, since it successfully retains identity information decipherable by machines. Moreover, even Forbes-G provides better results on the tasks of face verification on IJB-C, gender classification, and facial attribute classification than the existing algorithms.

Fig.~\ref{fig:qual_fv} compares obfuscated images in the LFW dataset qualitatively. We see that PRO-Face (Blur) retains some facial information recognizable by humans, although it achieves a higher accuracy on the LFW dataset than Forbes-G does. Similarly, the other versions of PRO-Face fail to achieve sufficient HI. Also, FALCO \cite{barattin2023attribute}, which is a face anonymization method, retains some facial attributes, lowering the HI. In contrast, both Forbes-G and Forbes-T make it challenging for humans to decipher the obfuscated images. More qualitative results on the other datasets are presented in Fig.~{\red S-1} \black in the supplement.

\begin{figure}[t]
    \centering
    \includegraphics[width=\linewidth]{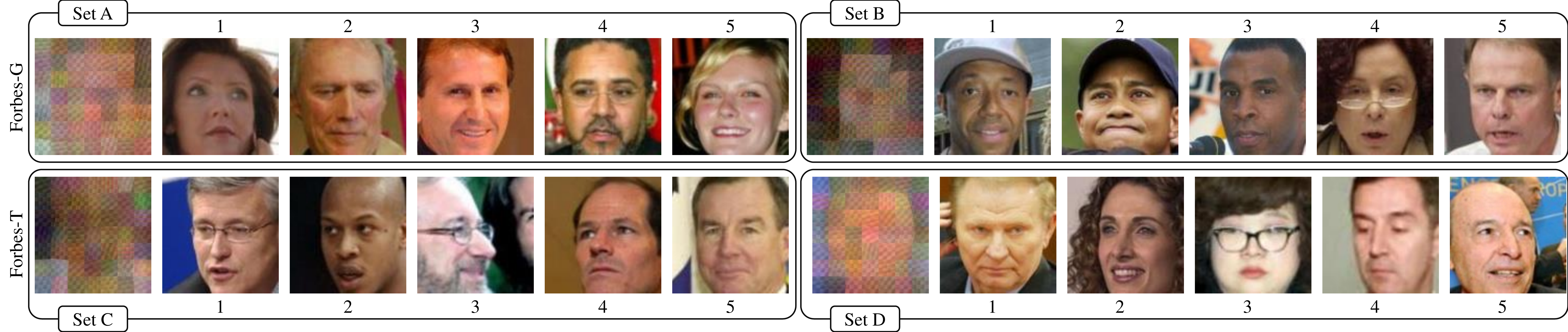}
    \caption{Examples of the questions in the user study. A subject was required to select one image, from the five candidates, containing the same identity as the obfuscated image.  The correct answers are {\scriptsize A) 3, B) 4, C) 2, D) 1}.}
    \label{fig:mos}
\end{figure}

\subsection{User Study}
\label{sec:user}

Forbes improves HI by minimizing the HI energies in \eqref{eq:margin_u} and \eqref{eq:margin_p}. To assess the HI of obfuscated images, we conducted a user study. We compared Forbes with the previous methods. Specifically, we used Blur, Pixelate, PRO-Face (Blur), PRO-Face (Pixelate), PRO-Face (FaceShifter), PRO-Face (SimSwap), FALCO, Forbes-G, and Forbes-T to generate 30 obfuscated images, respectively. Each obfuscated image was shown to a subject with five reference images: one with the same identity and the other four with different identities. The input image for generating the obfuscated image was different from the reference image of the same identity; they did not share the same background and poses.

We recruited 15 volunteers, who were asked to choose one reference image, which had the same identity as the obfuscated image, from the five candidates. We defined the correctness rate as the ratio of the right answers over the 30 tests. Then, we computed the average correctness rate (ACR) of the 15 volunteers.

Table \ref{tab:quantitative} compares these user study results on the LFW dataset. Since we provided multiple choice problems of five reference images, the ACR of random guess should be around 0.2. Note that, compared with Blur and Pixelate, the two PRO-Face versions worsen the ACR scores. In contrast, the proposed Forbes-T yields a low ACR of 0.2422, close to 0.2. This indicates that Forbes achieves high HI. Fig.~\ref{fig:mos} shows examples of the questions in this user study. More examples are presented in Fig.~{\red S-3} \black and {\red S-5} \black in the supplement.

\subsection{Analysis}
\label{sec:analysis}
We conduct ablation studies to analyze the main components of Forbes: block size, margins for the HI energies, energy functions, and obfuscating transformations.

\begin{figure}[t]
    \centering
    \includegraphics[width=\linewidth]{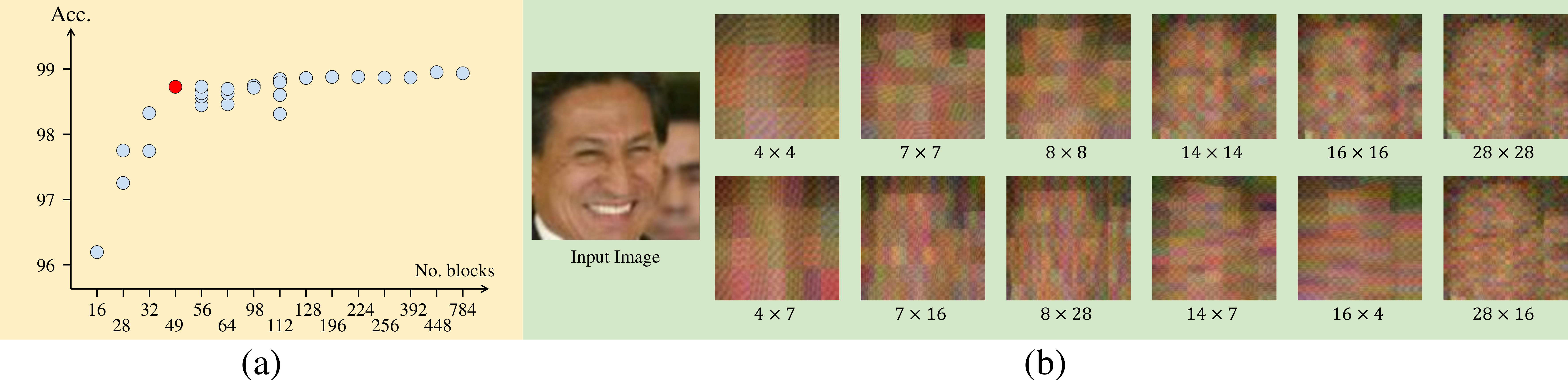}
    \caption{Impacts of the number $M\times N$ of blocks on obfuscation results. (a) Verification accuracies of Forbes-T on the LFW dataset according to $M\times N$. (b) Qualitative results. }
    \label{fig:blocksize}
\end{figure}

\begin{figure}[t]
    \centering
    \includegraphics[width=\linewidth]{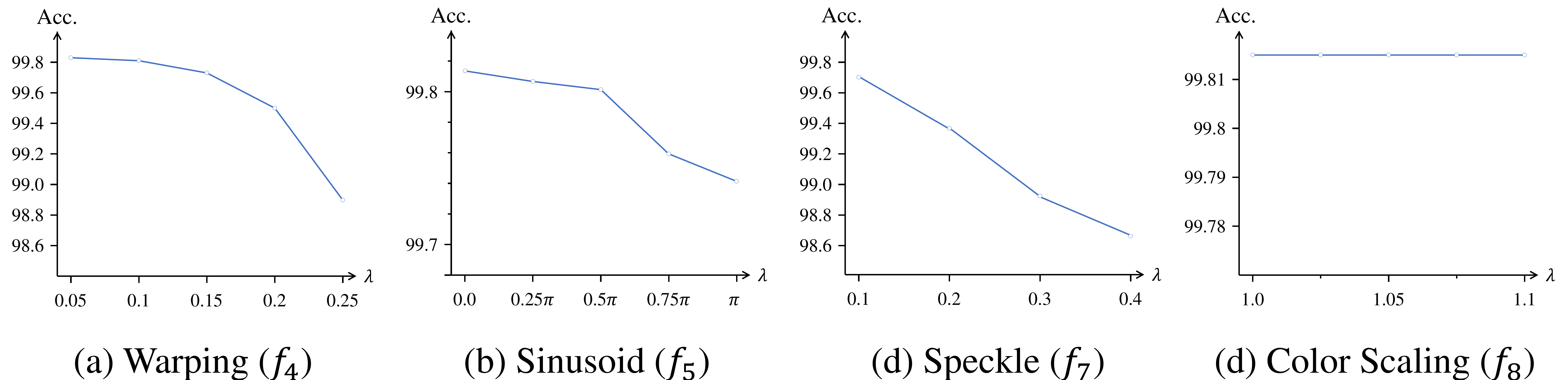}
    \caption{Verification accuracies of single obfuscating transformations $f_4$, $f_5$, $f_7$, and $f_8$ according to margins $\lambda$. }
    \label{fig:ab_margin}
\end{figure}

\textbf{Block size:} We analyze the impacts of the number $M\times N$ of blocks on obfuscation results in Fig.~\ref{fig:blocksize}. We test many block configurations from $4\times 4$ to $28\times 28$ and plot the verification accuracies of Forbes-T according to the number of blocks. In the graph, as more blocks are used, the verification accuracy improves in general. However, we see that, with more blocks, visually salient information remains in the obfuscated image, lowering HI. Hence, as a tradeoff between HI and MD, we use $7\times7$ blocks in the Forbes algorithm.

\textbf{Margins:}
We also analyze the impacts of the margins $\lambda$ for the HI energies in \eqref{eq:margin_u} and \eqref{eq:margin_p}. Fig.~\ref{fig:ab_margin} plots the verification accuracies of single obfuscating transformations $f_4$, $f_5$, $f_7$, and $f_8$ according to margin values. We see that, for $f_4$, $f_5$, and $f_7$, a larger margin corresponds to a lower machine accuracy. In other words, for MD, it is desirable to set a small margin. However, except $f_5$, the HI energies are designed to achieve a higher HI level with a larger margin. Hence, for HI, it is better to set a large margin. Note that, in $f_5$, the axis angles of the sinusoidal noise are used as parameters, which have no impact on HI. Therefore, by considering both MD and HI, we select margins $0.05$, $0.0$, $0.1$, and $1.05$ for $f_4$, $f_5$, $f_7$, and $f_8$, respectively, as listed in Table~\ref{table:transformation}.

\begin{table*}[t]
    \caption{Ablation studies for the energy functions on the LFW dataset \cite{huang2008lfw}. Verification accuracies of AdaFace \cite{kim2022adaface} are reported.}
    \centering
    \resizebox{0.85\linewidth}{!}{
    \begin{tabular}{L{2.3cm}C{1.7cm}C{1.7cm}C{1.7cm}C{1.7cm}|C{2.3cm}} \toprule
    Method     &$\mathcal{E}_{U}$& $\mathcal{E}_{C}$ & $\mathcal{E}_{D}$&$\mathcal{E}_{S}$& Forbes-T \\ \midrule
    \RomNum{1} &                 &                   &                  &                 & 56.30 \\
    \RomNum{2} &\checkmark       &                   &                  &                 & 54.49\\
    \RomNum{3} &                 & \checkmark        &                  &                 & 54.64\\
    \RomNum{4} &                 &                   & \checkmark       &                 & 98.90\\
    \RomNum{5} &                 &                   &                  & \checkmark      & 99.11\\
    \RomNum{6} &\checkmark       & \checkmark        &                  &                 & 52.18\\
    \RomNum{7} &\checkmark       &                   & \checkmark       &                 & 81.42\\
    \RomNum{8} &\checkmark       &                   &                  & \checkmark      & 87.67\\
    \RomNum{9} &                 & \checkmark        & \checkmark       &                 & 83.30\\
    \RomNum{10}&                 & \checkmark        &                  & \checkmark      & 89.98\\
    \RomNum{11}&                 &                   & \checkmark       & \checkmark      & 99.27\\
    \RomNum{12}&\checkmark       & \checkmark        & \checkmark       &                 & 83.14\\
    \RomNum{13}&\checkmark       & \checkmark        &                  & \checkmark      & 83.10\\
    \RomNum{14}&\checkmark       &                   & \checkmark       & \checkmark      & 93.12\\
    \RomNum{15}&                 & \checkmark        & \checkmark       & \checkmark      & 96.68\\
    \RomNum{16}&\checkmark       & \checkmark        & \checkmark       & \checkmark      & 98.70 \\\bottomrule
    \end{tabular}
    }
    \label{table:energy}
\end{table*}

\begin{figure}[t]
    \centering
    \includegraphics[width=\linewidth]{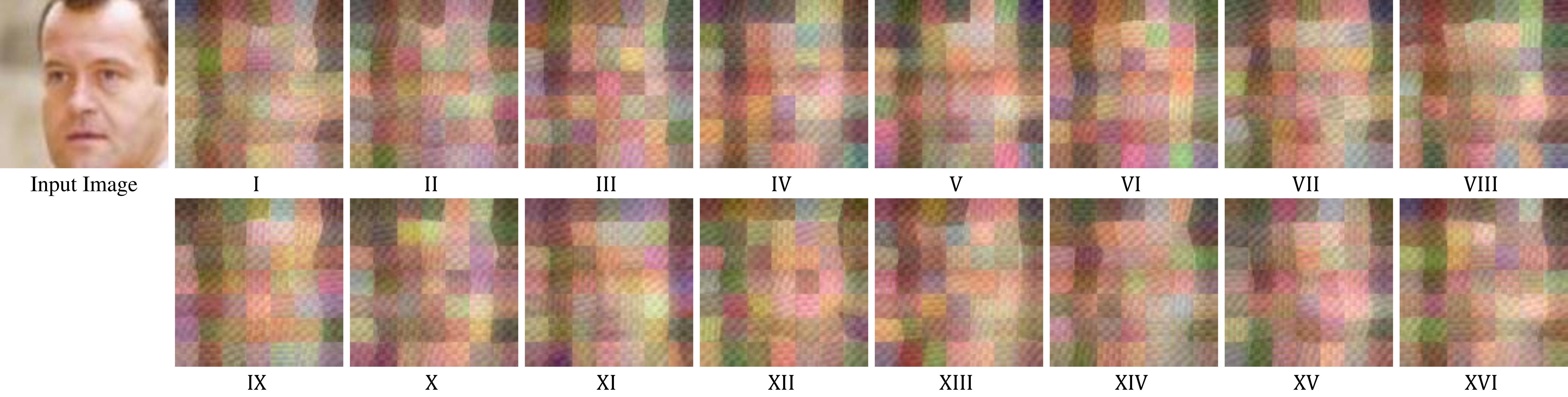}
    \caption{Examples of obfuscated images by the methods in Table~\ref{table:energy}.}
    \label{fig:energy}
\end{figure}

\textbf{Energy functions:}
Table \ref{table:energy} lists the verification accuracies on the LFW dataset according to combinations of the energy functions in the parameter optimization. Fig.~\ref{fig:energy} shows qualitative results of these combinations. First, method~\RomNum{1} does not use any energy functions; it uses randomly initialized parameters without optimization. Therefore, it yields the lowest accuracy. Without the HI energies $\mathcal{E}_{U}$ and $\mathcal{E}_{C}$, methods \RomNum{4}, \RomNum{5}, and \RomNum{11} can focus on MD only and thus yield slightly better accuracies than \RomNum{16}. However, in Fig.~\ref{fig:energy}, they yield lower levels of HI than \RomNum{16}. On the contrary, by comparing \RomNum{16} with the methods excluding the MD energies $\mathcal{E}_{D}$ and $\mathcal{E}_{S}$, we see that the verification accuracies degrade severely without the MD energies. In Fig.~\ref{fig:energy}, method \RomNum{1} and the methods without the MD energies have similar HI levels. Using all four energies, the proposed Forbes (\RomNum{16}) achieves a good balance between HI and MD.

\begin{table}[t!]
\begin{minipage}[h]{0.54\linewidth}
\renewcommand{\arraystretch}{1.1}
\caption{Ablation studies for the transformations on the LFW dataset \cite{huang2008lfw}. Verification accuracies of AdaFace \cite{kim2022adaface} are reported.}
\centering
\resizebox{\linewidth}{!}{
\begin{tabular}{L{1.5cm}C{1.5cm}C{1.5cm}C{1.5cm}C{1.5cm}|C{1.5cm}} \toprule
Method    &Averaging        & Warping           & Noising          & Scaling         & Forbes-T \\ \midrule
\RomNum{1}&\checkmark       &                   &                  &                 & 96.13    \\
\RomNum{2}&\checkmark       & \checkmark        &                  &                 & 96.91    \\
\RomNum{3}&\checkmark       & \checkmark        & \checkmark       &                 & 98.12    \\
\RomNum{4}&\checkmark       & \checkmark        & \checkmark       & \checkmark      & 98.70    \\ \bottomrule
\end{tabular}
}
\label{table:composite}
\end{minipage}\hfill
\begin{minipage}[h]{0.42\linewidth}
\centering
\includegraphics[width=5.0cm,height=2.2cm]{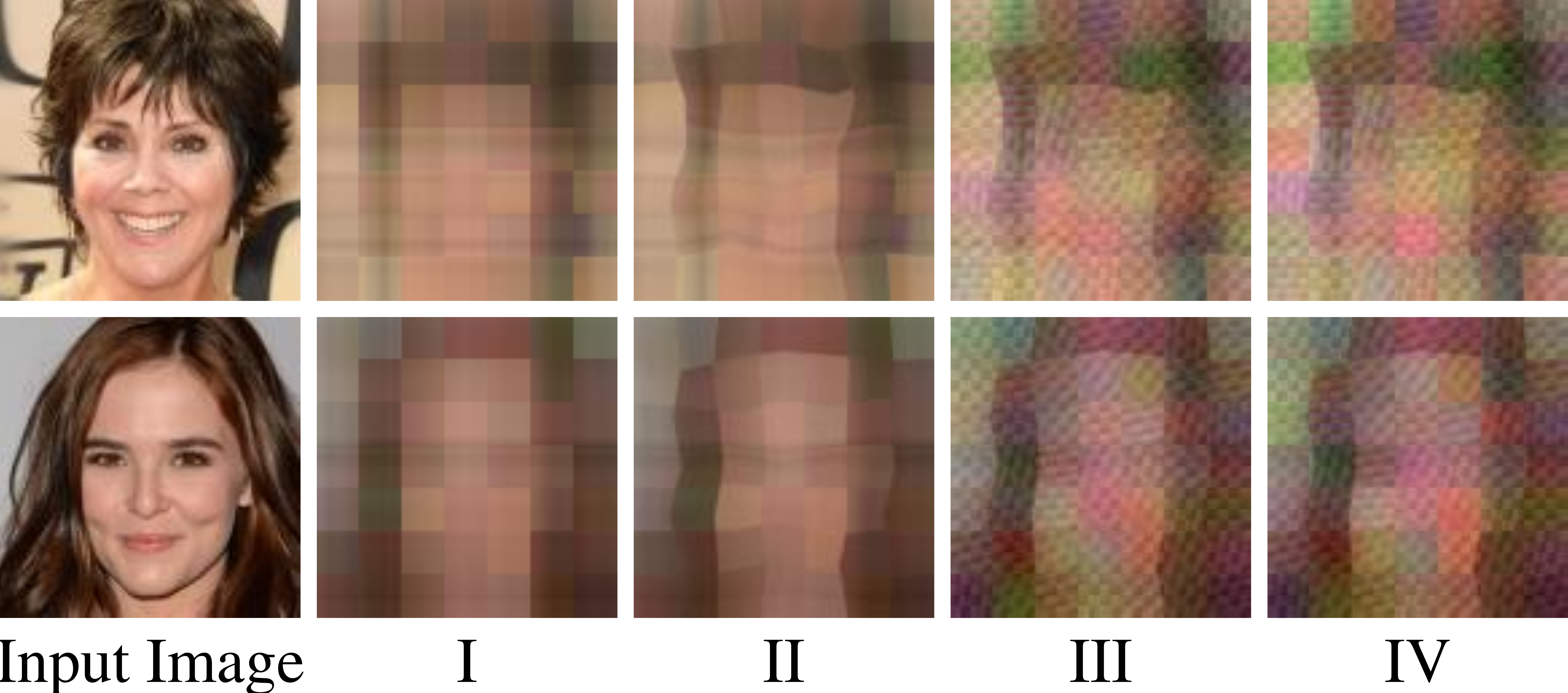}
\captionof{figure}{Examples of obfuscated images by the methods in Table~\ref{table:composite}.}
\label{fig:composite}
\end{minipage}
\end{table}

\textbf{Obfuscating transformations:}
We fix the parameter values of $f_1$, $f_2$, $f_3$, and $f_6$ to $1$ to yield a high HI level. In Table \ref{table:composite} and Fig.~\ref{fig:composite}, method \RomNum{1} effectively erases the high-frequency information while achieving a machine accuracy of 96.13\%. Method \RomNum{2}, which applies $f_4$ to the superposed image, warps the image and partly reconstructs high-frequency information. Note that method \RomNum{2} yields a slightly better accuracy than method \RomNum{1}. Since the warping cannot recover the information completely, method \RomNum{3} deceives the machine by adding a noise map to the warped image. Moreover, we adopt $f_8$ to scale each block and make it easy for machines to decipher, yielding the highest MD.

\section{Conclusions}
We proposed a novel algorithm for face obfuscation, called Forbes, to strike a good balance between HI and MD. To this end, we designed eight parameterized obfuscating transformations, combined them into a composite transformation, and developed an optimization scheme for the parameters based on BRS. Through extensive experiments on various facial datasets, it was demonstrated that the proposed Forbes achieves face obfuscation more effectively than the conventional algorithms.

\section*{Acknowledgements}
This work was supported by the 42dot Inc. and the National Research Foundation of Korea (NRF) grants funded by the Korea government (MSIT) (No.~NRF-2022R1A2B5B03002310 and No.~RS-2024-00397293).

\bibliographystyle{splncs04}
\bibliography{Forbes}

\clearpage

\renewcommand{\thesection}{\Alph{section}}

\renewcommand{\thetable}{S-\arabic{table}}
\renewcommand{\thefigure}{S-\arabic{figure}}
\begin{center}
    \Large \textbf{Forbes: Face Obfuscation Rendering via \\ Backpropagation Refinement Scheme} \\ \vspace*{0.4cm}
    \large \textbf{Supplemental Material}
\end{center}
\setcounter{section}{0}
\setcounter{figure}{0}

\section{Qualitative Results}

\begin{figure}
  \centering
  \resizebox{0.9\linewidth}{!}{
  \includegraphics[width=\linewidth]{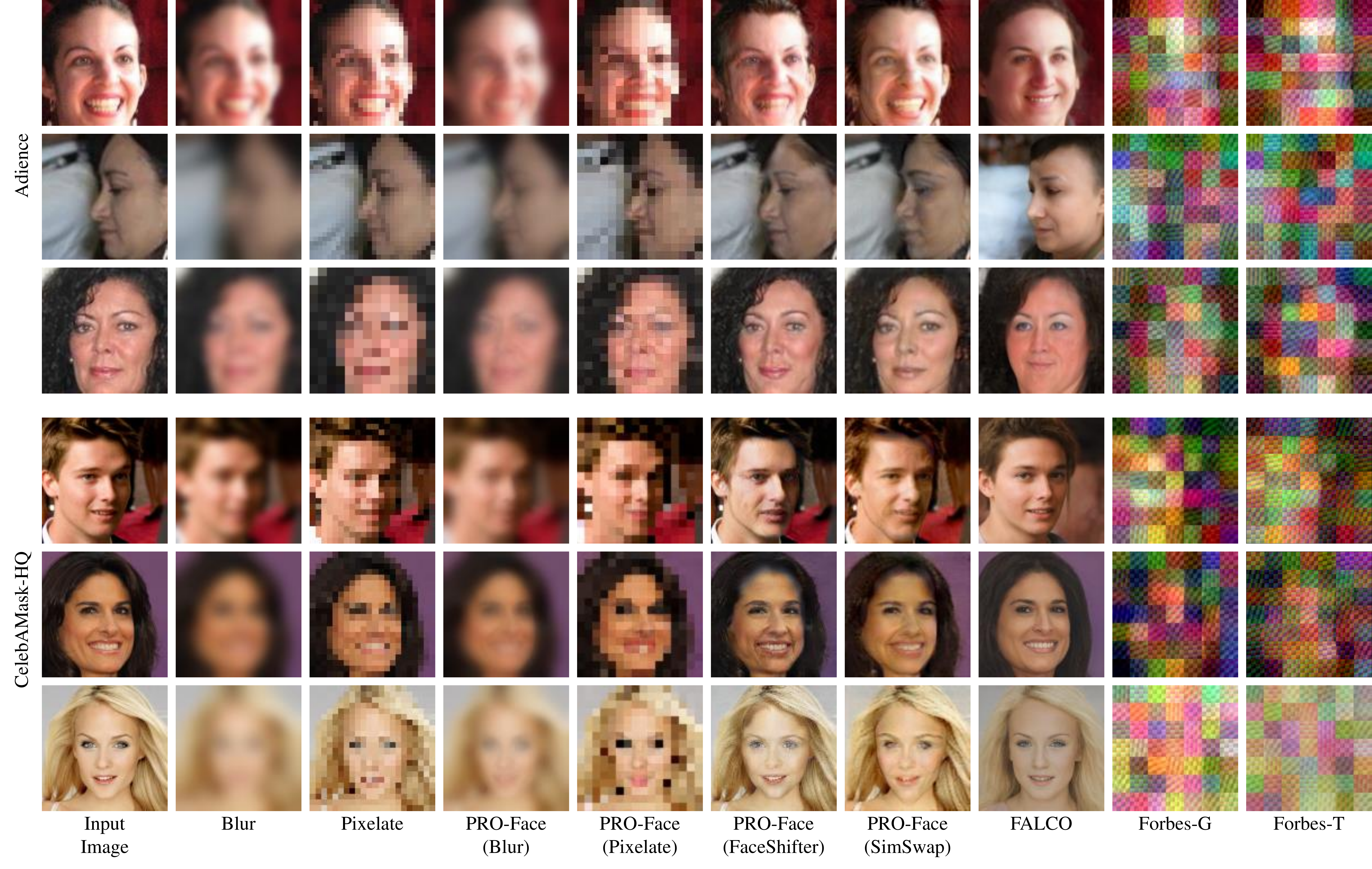}
  }
  \caption{Qualitative comparison of obfuscated images in the Adience and CelebA\-Mask-HQ datasets.}
  \label{fig:qual}
\end{figure}

Fig.~\ref{fig:qual} shows face obfuscation results on the Adience~\cite{eidinger2014adience} and CelebAMask-HQ~\cite{CelebAMask-HQ} datasets. Four versions of PRO-Face~\cite{yuan2022pro} (Blur, Pixelate, FaceShifter, and SimSwap) yield obfuscated images that still contain facial attribute information such as ages, genders, and hair colors, since they focus on obfuscating identity information only. Similarly, FALCO~\cite{barattin2023attribute} generates fake images that have the same facial attributes as the original images. Thus, its output can be easily recognized by humans. In contrast, the proposed Forbes algorithm removes successfully the attribute information, as well as the identity information, recognizable by humans.

\clearpage

\section{Failure Cases}

\begin{figure}
  \centering
  \resizebox{0.99\linewidth}{!}{
  \includegraphics[width=\linewidth]{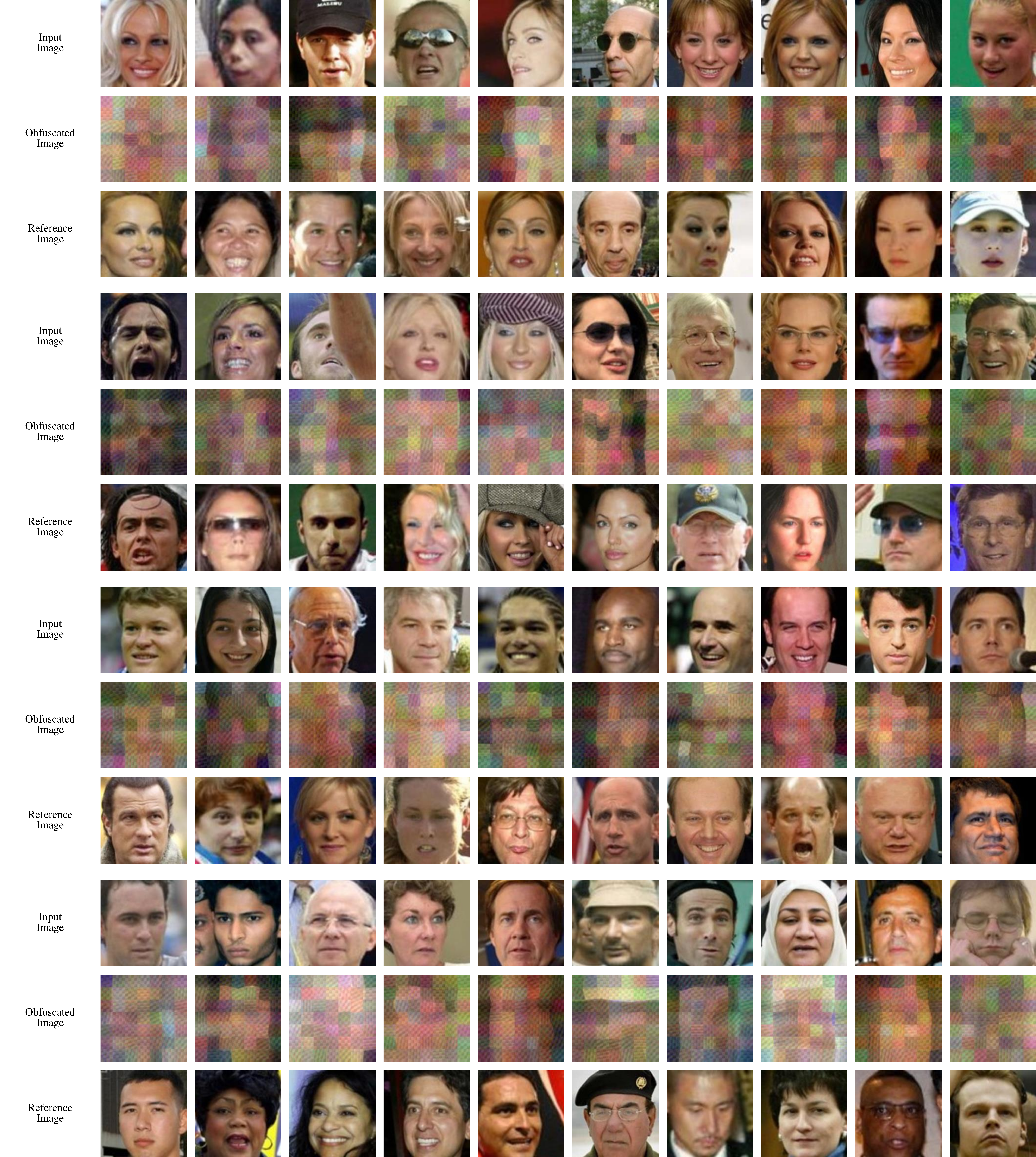}
  }
  \caption{Failure cases on the LFW dataset. The top two groups are for verifying the same identities, while the bottom two are for telling apart different identities.}
  \label{fig:failure}
\end{figure}

Fig.~\ref{fig:failure} shows failure cases on the LFW dataset~\cite{huang2008lfw}. The obfuscated images are unrecognizable by both humans and machines because it is difficult to match (or discriminate) even the original input and reference images.

\clearpage

\section{User Study}

\begin{figure}[h]
  \centering
  \resizebox{0.75\linewidth}{!}{
  \includegraphics[width=\linewidth]{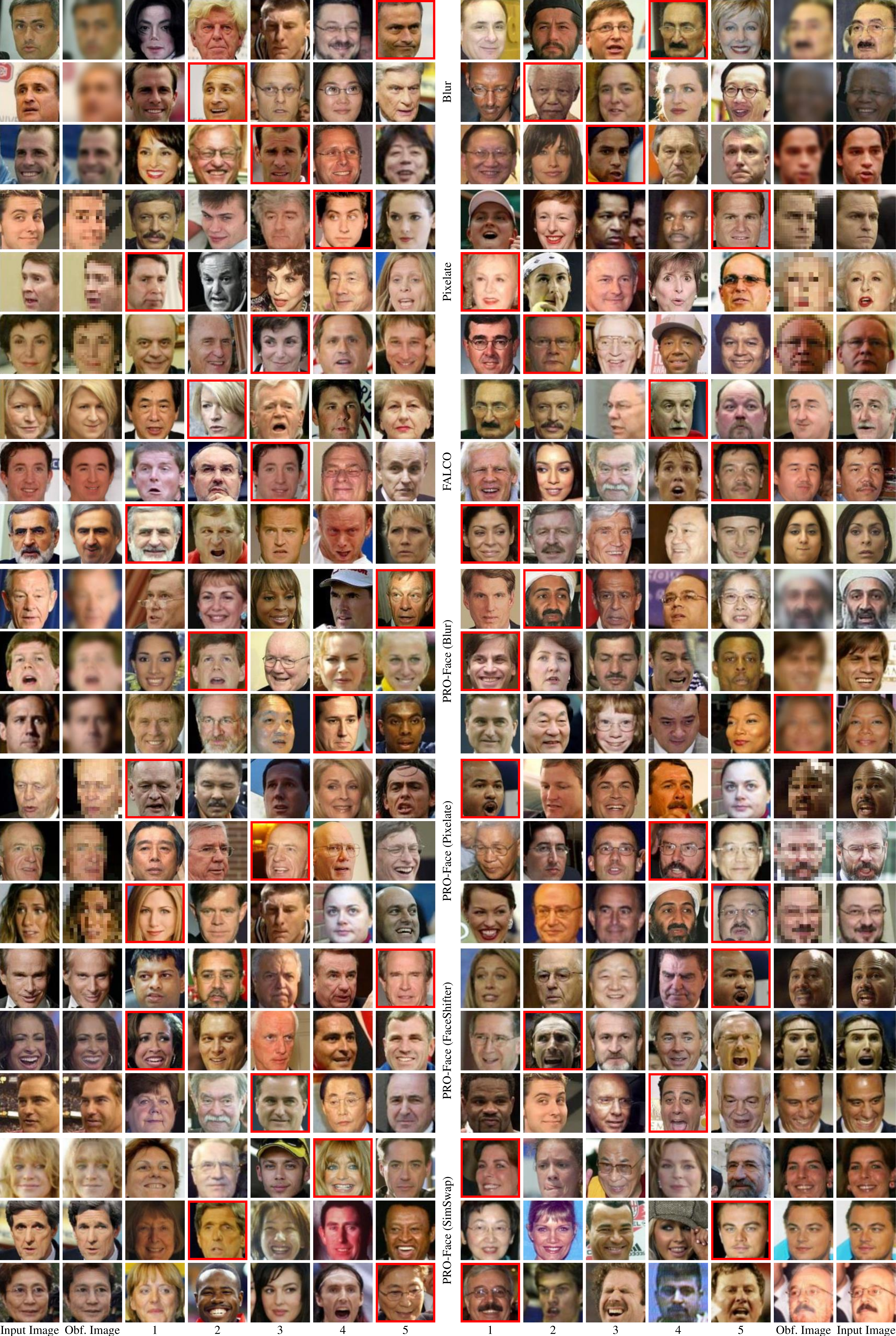}
  }
  \caption{Examples of the questions in the user study. The red boxes indicate the answers.}
  \label{fig:A1}
\end{figure}

\begin{figure}[h]
  \centering
  \resizebox{0.75\linewidth}{!}{
  \includegraphics[width=\linewidth]{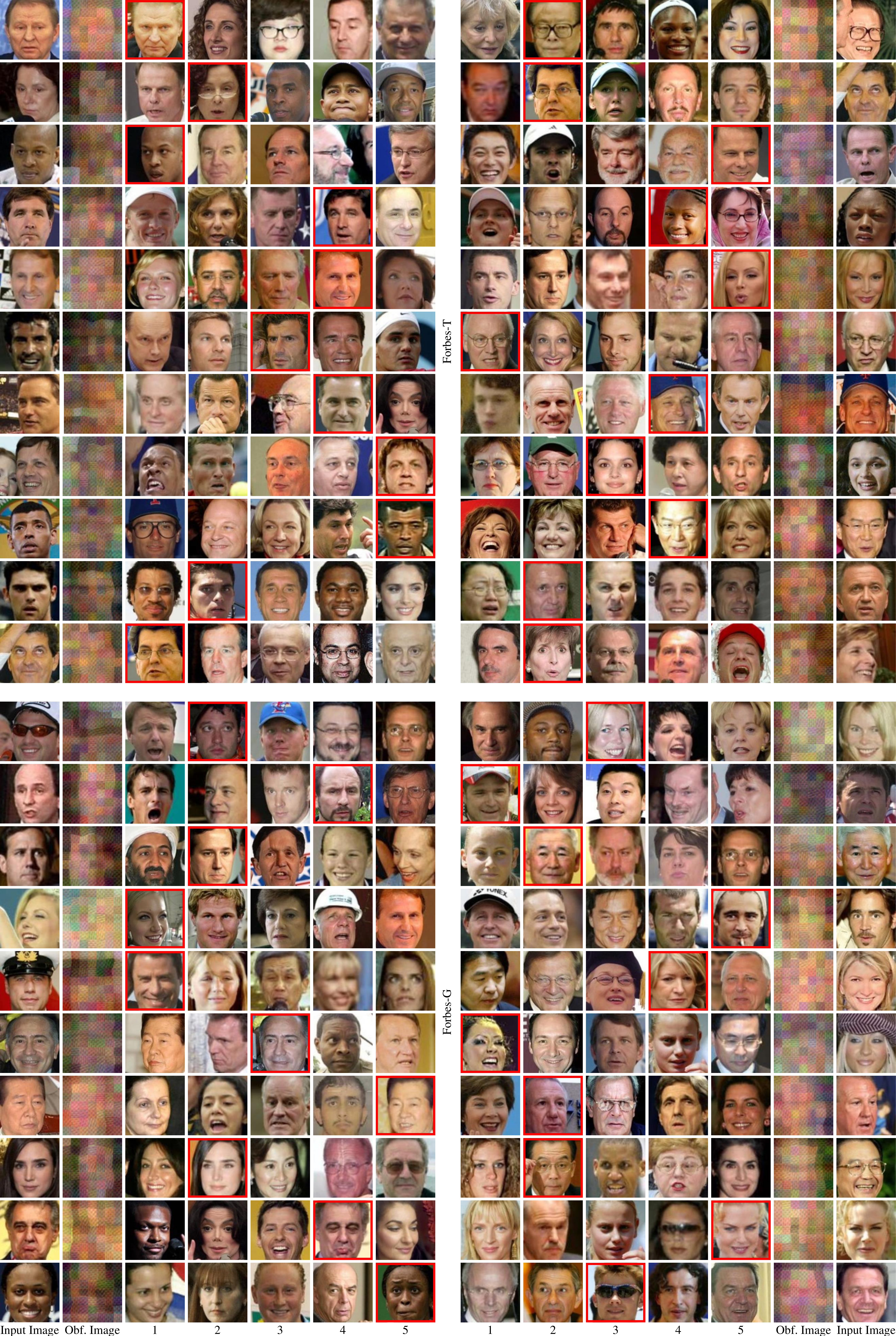}
  }
  \caption{Examples of the questions in the user study. The red boxes indicate the answers.}
  \label{fig:A2}
\end{figure}

Below in Fig.~\ref{fig:Q1} and Fig.~\ref{fig:Q2}, we provide more examples of the questions in the user study. The answers are marked on the next pages in Fig.~\ref{fig:A1} and Fig.~\ref{fig:A2}.

\begin{figure}[h]
  \centering
  \resizebox{0.75\linewidth}{!}{
  \includegraphics[width=\linewidth]{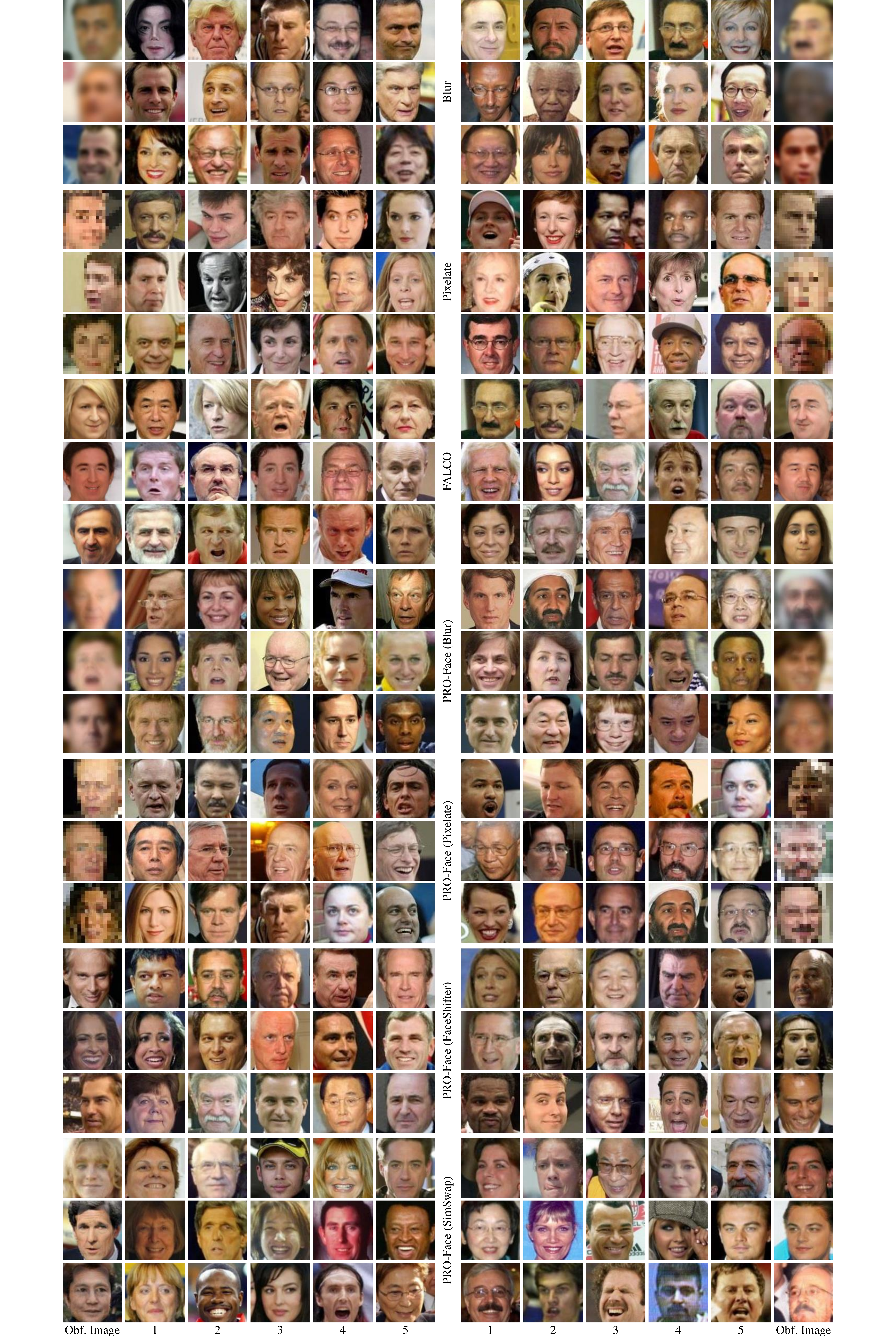}
  }
  \caption{Examples of the questions in the user study. The answers are on the next page.}
  \label{fig:Q1}
\end{figure}

\clearpage

\vspace*{1.334cm}
\clearpage

\clearpage

\vspace*{1.334cm}
\begin{figure}[h]
  \centering
  \resizebox{0.75\linewidth}{!}{
  \includegraphics[width=\linewidth]{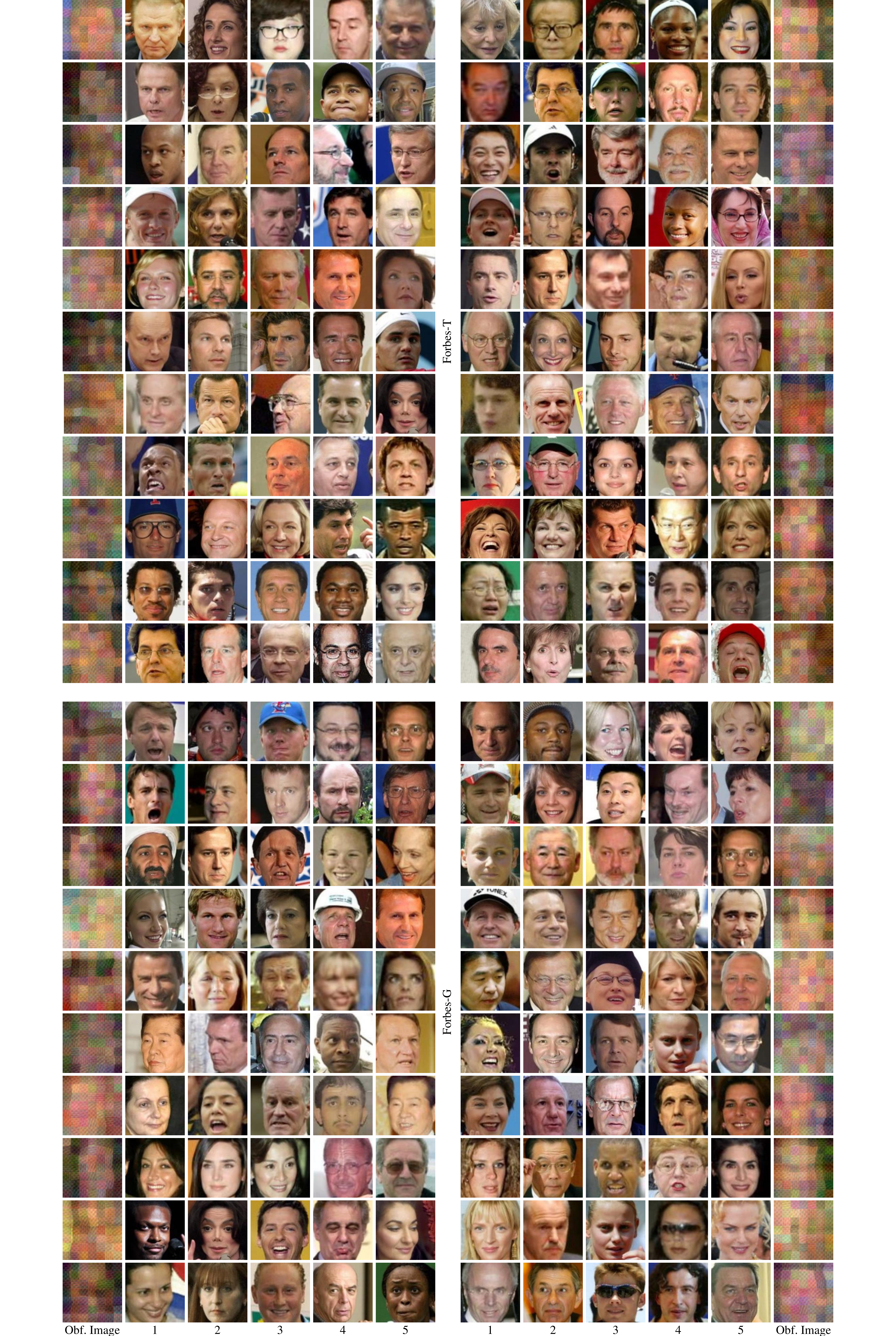}
  }
  \caption{Examples of the questions in the user study. The answers are on the next page.}
  \label{fig:Q2}
\end{figure}
\clearpage

\clearpage

\vspace*{1.334cm}
\clearpage

\end{document}